\documentclass[10pt,journal,compsoc]{IEEEtran}
\usepackage{amsmath,amsfonts}
\usepackage{algorithm}
\usepackage{algorithmic}
\usepackage{array}
\usepackage[caption=false,font=normalsize,labelfont=sf,textfont=sf]{subfig}
\usepackage{textcomp}
\usepackage{stfloats}
\usepackage{url}
\usepackage{verbatim}
\usepackage{graphicx}
\usepackage{cite}
\usepackage{bm}
\usepackage{enumitem} 
\usepackage{multirow}
\usepackage{booktabs}

\newtheorem{theorem}{Theorem}

\begin{document}

\title{Beyond Correlation: Causal Multi-View Unsupervised Feature Selection Learning}

\author{Zongxin Shen, Yanyong Huang, Bin Wang, Jinyuan Chang, Shiyu Liu, and Tianrui Li,~\IEEEmembership{Senior Member,~IEEE}
\thanks{Zongxin~Shen, Yanyong~Huang, Shiyu Liu and Jinyuan Chang are with the Joint Laboratory of Data Science and Business Intelligence, School of Statistics and Data Science, Southwestern University of Finance and Economics, Chengdu 611130, China (e-mail: zxshen@smail.swufe.edu.cn; huangyy@swufe.edu.cn; shyu.liu@foxmail.com; changjinyuan@swufe.edu.cn), Yanyong Huang is the corresponding author;}

\thanks{Bin Wang is with the College of Information Science and Engineering, Ocean University of China, Qingdao 266100, China (e-mail: wangbin9545@ouc.edu.cn);}

\thanks{Tianrui Li is with the School of Computing and Artificial Intelligence, Southwest Jiaotong University, Chengdu 611756, China (e-mail: trli@swjtu.edu.cn).}

\thanks{This work has been submitted to the IEEE for possible publication. Copyright may be transferred without notice, after which this version may no longer be accessible.}}

\markboth{Journal of \LaTeX\ Class Files,~Vol.~14, No.~8, August~2021}%
{Shell \MakeLowercase{\textit{et al.}}: A Sample Article Using IEEEtran.cls for IEEE Journals}

\IEEEpubid{0000--0000/00\$00.00~\copyright~2021 IEEE}

\maketitle

\begin{abstract}
Multi-view unsupervised feature selection (MUFS) has recently received increasing attention for its promising ability in dimensionality reduction on multi-view unlabeled data. Existing MUFS methods typically select discriminative features by capturing correlations between features and clustering labels. However, an important yet underexplored question remains: \textit{Are such correlations sufficiently reliable to guide feature selection?} In this paper, we analyze MUFS from a causal perspective by introducing a novel structural causal model, which reveals that existing methods may select irrelevant features because they overlook spurious correlations caused by confounders. Building on this causal perspective, we propose a novel MUFS method called CAusal multi-view Unsupervised feature Selection leArning (CAUSA). Specifically, we first employ a generalized unsupervised spectral regression model that identifies informative features by capturing dependencies between features and consensus clustering labels. We then introduce a causal regularization module that can adaptively separate confounders from multi-view data and simultaneously learn view-shared sample weights to balance confounder distributions, thereby mitigating spurious correlations. Thereafter, integrating both into a unified learning framework enables  CAUSA to select causally informative features. Comprehensive experiments demonstrate that CAUSA outperforms several state-of-the-art methods. To our knowledge, this is the first in-depth study of causal multi-view feature selection in the unsupervised setting.
\end{abstract}


\section{Introduction}
\IEEEPARstart{W}{ith} the widespread availability of high-dimensional, unlabeled multi-view data in practical applications, multi-view unsupervised feature selection (MUFS) has attracted increasing attention in recent years~\cite{li2024exploring,zhang2024scalable,zhang2024efficient,huang2025time}. It aims to identify representative features from multiple heterogeneous feature sets, thus effectively addressing the ``curse of dimensionality'' problem and improving the performance of downstream tasks. Existing MUFS methods typically leverage the cross-view local structures to learn clustering labels. Simultaneously, they capture correlations between features and the inferred clustering labels to guide feature selection, which has shown promising performance in practice~\cite{ CvLPDCL,UKMFS}.


Generally, the key question is \textit{whether the correlations captured by existing MUFS methods are sufficiently reliable to guide feature selection. If not, what factors undermine this reliability, and how can it be improved?} Investigating this question can deepen our understanding of the MUFS problem, reveal the limitations of existing methods, and inspire more effective solutions. The causal relationships are widely acknowledged to be among the most reliable and interpretable foundations for downstream tasks~\cite{feuerriegel2024causal}. For example, in animal image clustering, causal features (e.g., facial structure) inherently distinguish dogs from other animals, and selecting these features can therefore improve clustering performance. In contrast, non-causal features (e.g., background elements) do not provide such discriminative information, and their inclusion may degrade clustering performance. This observation motivates us to analyze MUFS from a causal perspective. To this end, we introduce a novel structural causal model (SCM) in Section 2. The SCM demonstrates that confounding factors that influence both features and clustering labels can induce spurious correlations. For instance, background features may be spuriously correlated with clustering labels if the dataset is biased towards specific environments (e.g., cats are often photographed indoors, while dogs are frequently photographed on grass). Such distribution bias may induce potential confounding, leading to spurious correlations between grass features and clustering labels~\cite{kuang2021balance,huang2025debiasing}. Since existing MUFS methods ignore such confounding effects, they often rely on spurious correlations to select non-causal features, such as grass features in the aforementioned example.

\begin{figure*}[t]
\centering
\includegraphics[width=\textwidth]{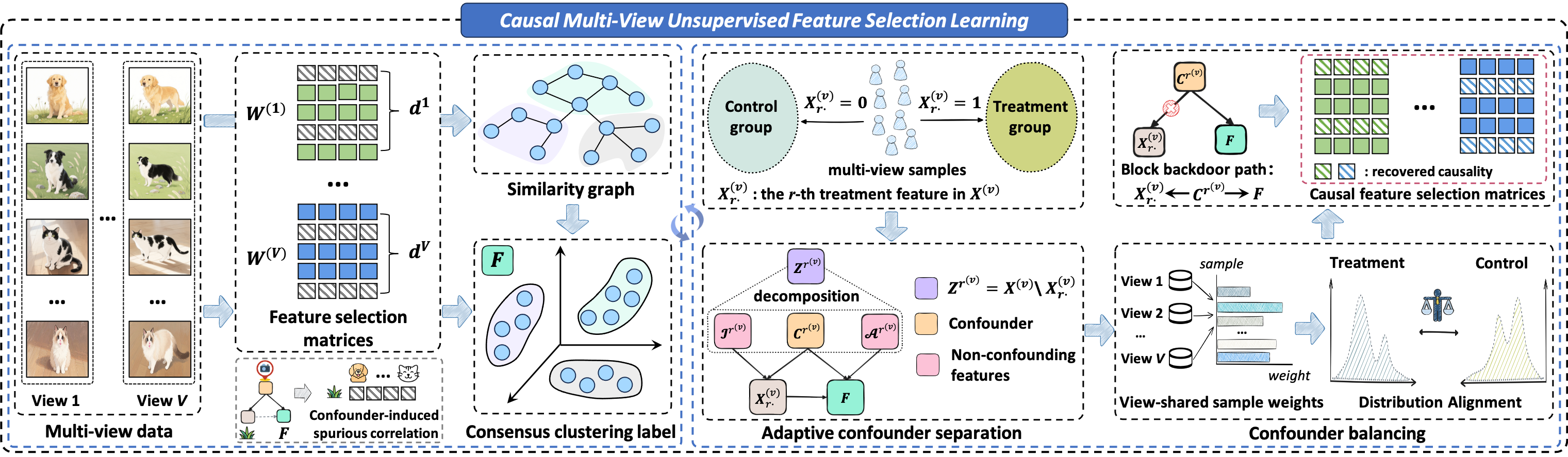} 
\caption{The framework of the proposed CAUSA.}
\label{Framework}
\end{figure*} 
Recognizing this limitation naturally raises the question of \textit{how to mitigate such spurious correlations and identify causally informative features}. Addressing this challenge in unsupervised settings is particularly difficult due to the lack of label information. Although several causal feature selection methods have been proposed recently, their reliance on label information makes them inapplicable to unsupervised scenarios~\cite{ling2022light,quinzan2023drcfs,mangal2025ate}. Based on Pearl's back-door criterion, spurious correlations can be mitigated through confounder adjustment~\cite{pearl2014confounding}. Accordingly, several confounder balancing methods have been proposed in the causal literature, which construct sample weights to balance confounder distributions, thereby blocking backdoor paths and reducing spurious correlations~\cite{shen2018causally,zubizarreta2015stable}. However, most existing confounder balancing methods treat all observed features as confounders, neglecting non-confounding features that only affect the clustering labels or the treatment features. Balancing these non-confounders may result in inaccurate causality identification by existing methods~\cite{liu2023debiased}. Therefore, how to adaptively separate confounding factors and mitigate spurious correlations in order to identify causally informative features from unlabeled multi-view data remains an urgent challenge in practical applications.

To address the above issues, we propose a novel MUFS method, called CAusal multi-view Unsupervised feature Selection leArning (CAUSA). Specifically, we first incorporate MUFS into a generalized unsupervised spectral regression model, which identifies informative features by capturing dependencies between features and consensus clustering labels. Then, we propose a novel causal regularization module that adaptively separates confounders from multiple views and simultaneously learns shared sample weights across views to balance confounder distributions, thus mitigating spurious correlations. These components are integrated into a unified learning framework, enabling CAUSA to select causally informative features. The overall framework of CAUSA is illustrated in Fig. \ref{Framework}. The main contributions of this paper are summarized as follows:
\begin{itemize}
    \item To the best of our knowledge, this is the first work to propose and study causal feature selection for unlabeled multi-view data, which advances MUFS beyond correlation toward a more reliable causal paradigm.
    \item We introduce a novel SCM to analyze MUFS from a causal perspective, revealing that existing methods are often misled by confounder-induced spurious correlations and consequently select causally irrelevant features.   
    \item By integrating a novel causal regularization module with MUFS into a unified learning framework, the proposed method adaptively separates confounders and learns view-shared weights to balance their distributions, thereby mitigating spurious correlations and facilitating the identification of causally informative features.
    \item An efficient alternative optimization algorithm is developed to solve the proposed CAUSA, and extensive experiments demonstrate the superiority of CAUSA over several state-of-the-art (SOTA) methods.
\end{itemize}

\section{Causal Analysis on MUFS}
\subsection{Notations and Problem Definition} 
Throughout the paper, matrices and vectors are written in bold uppercase letters, while scalars are written in regular font. Given any matrix $\bm{M} \in \mathbb{R}^{p \times q}$, its $(i,j)$-th entry, $i$-th row, and $j$-th column are denoted by $\mathnormal{M}_{ij}$, $\bm{M}_{i \cdot}$, and $\bm{M}_{\cdot j}$, respectively. The trace and transpose of $\bm{M}$ are respectively represented by $\operatorname{Tr}(\bm{M})$ and $\bm{M}^{\top}$. The Frobenius norm and $\ell_{2,1}$-norm of $\bm{M}$ are defined as $\|\bm{M}\|_{\mathrm{F}}=\sqrt{\sum_{i=1}^{p}\sum_{j=1}^{q}\mathnormal{M}_{ij}^{2}}$ and $\|\bm{M}\|_{2,1}=\sum_{i=1}^{p}\sqrt{\sum_{j=1}^{q}\mathnormal{M}_{ij}^{2}}$, respectively. $\bm{I}$ denotes the identity matrix, and $\textbf{1} = [1, \dots, 1]^{\top}$ represents a column vector of all ones.

Given an unlabeled multi-view dataset $\mathcal{X}=\{\bm{X}^{v} \in \mathbb{R}^{d_v \times n}\}_{v=1}^{V}$ with $V$ views, where $\bm{X}^{v}$ is the data matrix of the $v$-th view, consisting of $n$ samples and $d_v$ features. Our goal is to select causally informative features from the multi-view dataset $\mathcal{X}$ in an unsupervised manner. 

\subsection{Causal Perspective on MUFS} 
To analyze MUFS from a causal perspective, we introduce a structural causal model (SCM) based on the following assumptions: (\romannumeral1) The original feature set in $\bm{X}^{v} (v=1,\dots,V)$ can be partitioned into two disjoint subsets: causal features $\mathcal{R}^{(v)}$ and non-causal features $\mathcal{U}^{(v)}$, such that $\bm{X}^{v} = \mathcal{R}^{(v)} \cup \mathcal{U}^{(v)}$ and $\mathcal{R}^{(v)} \cap \mathcal{U}^{(v)} = \emptyset$. (\romannumeral2) Clustering serves as a representative downstream task in the unsupervised setting considered in this paper. Let $\mathcal{F}$ denote the clustering labels. $\mathcal{F}$ depends solely on causal features from all views, while non-causal features do not directly affect  $\mathcal{F}$. Formally, this implies that $P(\mathcal{F}|\{\mathcal{R}^{(v)},\mathcal{U}^{(v)}\}_{v=1}^{V})=P(\mathcal{F}|\{\mathcal{R}^{(v)}\}_{v=1}^{V})$. (\romannumeral3) There are potential confounders $\mathcal{C}^{(v)}$ in each view that simultaneously influence both the non-causal features $\mathcal{U}^{(v)}$ and clustering labels $\mathcal{F}$. 
 
\begin{figure}[t]
\centering
\includegraphics[width=0.8\columnwidth]{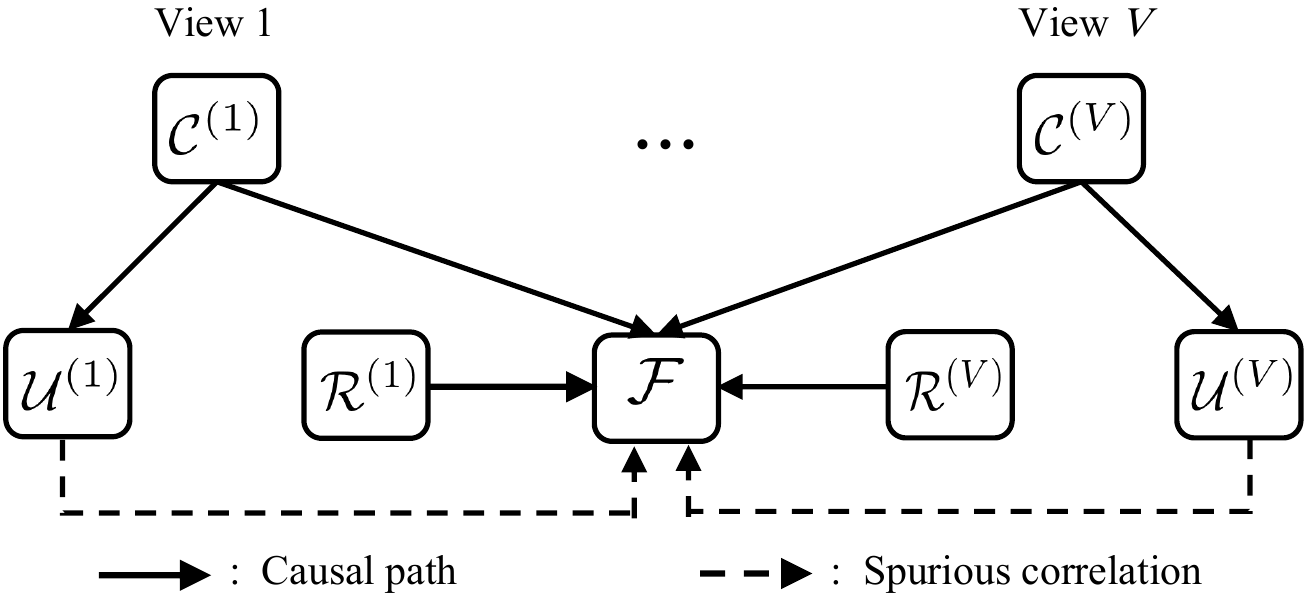} 
\caption{The SCM graph for MUFS.}
\label{SCM}
\end{figure}

Based on the above assumptions, the SCM can be illustrated by the causal graph shown in Fig.~\ref{SCM}. To provide an intuitive understanding, consider the task of clustering animal images taken from different camera angles, where each angle represents a distinct view. Within each view, features such as fur texture and facial structure are causal features that fundamentally determine cluster assignment of the animals, i.e., $\mathcal{R}^{(v)} \rightarrow \mathcal{F}$. In contrast, background elements like grass or flooring are non-causal features that do not influence cluster assignment and therefore have no direct causal path to $\mathcal{F}$. However, as previously discussed, sample distribution bias may introduce potential confounding, thereby resulting in spurious correlations between non-causal features and $\mathcal{F}$ via the backdoor path $\mathcal{U}^{(v)} \leftarrow \mathcal{C}^{(v)} \rightarrow \mathcal{F}$. To demonstrate this spurious correlation, we can derive the joint distribution of $\mathcal{F}$ and $\mathcal{U}^{(v)}$ from the SCM as follows:
\begin{equation}\label{I.1}
P(\mathcal{F}, \mathcal{U}^{(v)})\!=\!\!\sum_{\mathcal{C}^{(v)}} P(\mathcal{F}\!\mid\!\mathcal{C}^{(v)}) P(\mathcal{U}^{(v)} \!\mid\! \mathcal{C}^{(v)}) P(\mathcal{C}^{(v)}).
\end{equation}

In Eq. (\ref{I.1}), the joint distribution $P(\mathcal{F}, \mathcal{U}^{(v)})$ cannot be factorized into the product $P(\mathcal{F})P(\mathcal{U}^{(v)})$ by marginalization over $\mathcal{C}^{(v)}$, since $\mathcal{C}^{(v)}$ influences both clustering labels $\mathcal{F}$ and non-causal features $\mathcal{U}^{(v)}$. This indicates that $\mathcal{C}^{(v)}$ induces a statistical dependence between $\mathcal{F}$ and $\mathcal{U}^{(v)}$, even though there is no direct causal relationship between them. Such dependence gives rise to spurious correlations. If these confounding effects are ignored, traditional MUFS methods may mistakenly select non-causal features $\mathcal{U}^{(v)}$ as important features, which in turn degrades the performance of the downstream clustering task. Therefore, it is crucial to address confounding-induced spurious correlations to identify causally informative features.

\section{The Proposed Method}\label{Sec:The Proposed Method}
In this section, we provide a detailed introduction to our causality-inspired MUFS method, CAUSA. We begin by employing a generalized unsupervised spectral regression model that performs multi-view feature selection by capturing dependencies between features and consensus clustering labels across different views. Building on this, we develop a novel causal regularization module that adaptively separates confounders from multi-view data and learns view-shared sample weights to balance confounder distributions. Finally, CAUSA integrates these components into a unified learning framework, simultaneously mitigating spurious correlations and identifying causally informative features.

\subsection{Formulation of CAUSA} 
Unlike existing causal feature selection methods that rely on label information, we propose an unsupervised approach that selects causally informative features by learning each feature’s causal contribution to cluster separation. Specifically, we first use a generalized unsupervised spectral regression model on multi-view data to identify informative features by capturing dependencies between features and consensus clustering labels. It can be formulated as follows:
\begin{equation}\label{M.1}
\begin{aligned}
&\min_{\bm{W}^{\!(v)}\!,\bm{F}}\!\sum_{v=1}^{V} \!\alpha\|\bm{X}^{(v) \top}\!\bm{W}^{(v)} \!\!-\!\! \bm{F} \|_{\mathrm{F}}^{2}\!+\!\lambda \|\bm{W}^{(v)}\|_{2,1}\!\!+\!\!\operatorname{Tr}(\bm{F}^{\!\top}\!\!\bm{L}\bm{F})\\
&\text { s.t. } \bm{F}^{\top}\bm{F}=\bm{I}, \bm{F} \geq 0, \bm{W}^{(v) \top}\bm{W}^{(v)}=\bm{I},
\end{aligned}
\end{equation}
where $\bm{W}^{(v)} \in \mathbb{R}^{d_v \times c}$ denotes the feature selection matrix of the $v$-th view, $\bm{F} \in \mathbb{R}^{n \times c}$ represents the consensus clustering label across different views, and $c$ is the dimension of the low-dimensional space. Moreover, $\bm{L}$ denotes the cross-view Laplacian matrix, constructed using the method proposed in~\cite{zhou2006learning}. In Eq.~(\ref{M.1}), the $i$-th row of $\bm{W}^{(v)}$ captures the dependence between the feature $\bm{X}_{i \cdot}^{(v)}$ and the clustering label $\bm{F}$, where a stronger dependence indicates a greater contribution of  $\bm{X}_{i \cdot}^{(v)}$ to distinguishing clusters. Accordingly, $\|\bm{W}_{i \cdot}^{(v)}\|_{2}^{2}$ can serve as a quantitative measure of the importance of the $i$-th feature.

Due to confounding bias in real-world multi-view data, $\bm{W}^{(v)}$ may capture spurious correlations and consequently select irrelevant features, as previously discussed. Therefore, mitigating these spurious correlations is essential for identifying causally informative features.

\begin{theorem}\label{T1}
Suppose the confounder $\mathcal{C}^{(v)}$ is properly adjusted so that, under the adjusted distribution $\tilde{P}$, it is independent of the non-causal features $\mathcal{U}^{(v)}$, i.e., $\tilde{P}(\mathcal{C}^{(v)} \mid \mathcal{U}^{(v)} = u^{(v)}) = \tilde{P}(\mathcal{C}^{(v)})$ for all $u^{(v)}$. Then, $\mathcal{U}^{(v)}$ and $\mathcal{F}$ are independent in SCM, as shown by 
\begin{equation}
\begin{aligned}
\tilde{P}(\mathcal{F}, \mathcal{U}^{(v)}) &= \tilde{P}(\mathcal{U}^{(v)}) \sum_{\mathcal{C}^{(v)}} \tilde{P}(\mathcal{F} \mid \mathcal{C}^{(v)}) \tilde{P}(\mathcal{C}^{(v)}) \\
&= \tilde{P}(\mathcal{U}^{(v)}) \tilde{P}(\mathcal{F}).
\end{aligned}
\end{equation}
\end{theorem}

In Theorem 1, we demonstrate that spurious correlations between non-causal features and clustering labels can be eliminated by ensuring that the confounder distribution remains consistent across different values of the non-causal features. This is consistent with Pearl’s backdoor criterion~\cite{pearl2014confounding}, which asserts that adjusting for confounders (i.e., ensuring $\mathcal{C}^{(v)} \perp \mathcal{U}^{(v)}$ in the adjusted distribution) blocks spurious paths and enables causal identification. Motivated by this observation, we propose a novel causal regularization module that learns sample weights to balance the distribution of confounders, which are adaptively identified from multiple views. 

Specifically, given a treatment feature $\bm{X}_{r \cdot}^{(v)}$ in view $v$, we consider the remaining features that are associated with both $\bm{X}_{r \cdot}^{(v)}$ and the clustering label as confounders, denoted by $\bm{C}^{r^{(v)}}$. We identify these confounders using our proposed adaptive confounder separation method, which will be introduced later. Then, the samples characterized by confounders $\bm{C}^{r^{(v)}}$ are partitioned into the treatment and control groups according to values of the treatment feature $\bm{X}_{r \cdot}^{(v)}$. Without loss of generality, we define the treatment feature as a binary variable taking values of 0 or 1, assigning samples to the treatment group when the value is 1 and to the control group when it is 0. To mitigate confounding bias and capture true causality of $\bm{X}_{r \cdot}^{(v)}$, we learn a set of sample weights to balance the confounder distribution between treatment and control groups. Meanwhile, we employ Maximum Mean Discrepancy (MMD)~\cite{gretton2006kernel} to quantify distributional differences, as this kernel-based metric characterizes complex nonlinear relationships by embedding distributions into a Reproducing Kernel Hilbert Space (RKHS). The corresponding objective function is formulated below:
\begin{equation}\label{M.2}
\begin{aligned}
&\min_{\bm{\tau}^{(v)}}\! \| \!\frac{1}{|\nabla_{r}^{(v)}\!|}\!\!\sum_{i\in \nabla_{r}^{(v)}}\!\!\!\!\phi(\tau_{i}^{(v)} \!\bm{C}_{\cdot i}^{r^{(v)}}\!) \!-\! \frac{1}{|\Delta_{r}^{(v)}\!|}\!\!\sum_{j\in\Delta_{r}^{(v)}}\!\!\!\!\phi(\tau_{j}^{(v)}\!\bm{C}_{\cdot j}^{r^{(v)}}\!)\|_{\mathcal{H}_{\kappa}}^{2}\\
&\text { s.t. } \tau_{i}^{(v)} \geq 0, \bm{1}^{\top}\bm{\tau}^{(v)}=1,
\end{aligned}
\end{equation}
where $\bm{\tau}^{(v)} \in \mathbb{R}^{n \times 1}$ denotes the vector of sample weights in the $v$-th view. The sets $\nabla_{r}^{(v)}$ and $\Delta_{r}^{(v)}$ indicate the indices of samples in the treatment and control groups, respectively, with $|\nabla_{r}^{(v)}|$ denoting the cardinality of $\nabla_{r}^{(v)}$. Moreover, $\phi(\tau_{i}^{(v)} \!\bm{C}_{\cdot i}^{r^{(v)}}) = \kappa(\tau_{i}^{(v)} \!\bm{C}_{\cdot i}^{r^{(v)}}, \cdot)$ represents the feature map induced by the kernel $\kappa(\cdot, \cdot)$, and $\| \cdot\|_{\mathcal{H}_{\kappa}}$ denotes the RKHS norm~\cite{cui2020calibrated}. 

To enhance the balancing process in Eq.~(\ref{M.2}), we propose an adaptive confounder separation method that, unlike conventional causal approaches, does not treat all observed features as potential confounders. Specifically, for the treatment feature $\bm{X}_{r \cdot}^{(v)}$, the remaining features, denoted as $\bm{Z}^{r^{(v)}}\!\!\!=\!\bm{X}^{(v)} \backslash \bm{X}_{r \cdot}^{(v)} \!\in\! \mathbb{R}^{(d_v-1) \times n}$, can typically be divided into three subsets: confounders, adjustment variables, and instrumental variables~\cite{wu2022learning}. Due to the lack of prior knowledge about the true causal relationships, we propose a data-driven approach to identify confounders from $\bm{Z}^{r^{(v)}}$ by exploiting their dual associations with both treatment feature and clustering label. More specifically, we introduce a learnable confounding indicator vector $\bm{e}^{r^{(v)}} \in \{0,1\}^{(d_v-1) \times 1} $, where ${e}_{i}^{r^{(v)}}=1$ if the $i$-th feature of $\bm{Z}^{r^{(v)}}$ is a confounder, and ${e}_{i}^{r^{(v)}}=0$ otherwise. Thus, $\bm{C}^{r^{(v)}}$ can be written as $\bm{C}^{r^{(v)}}= (\bm{e}^{r^{(v)}}\textbf{1}^{\top}) \odot \bm{Z}^{r^{(v)}}$, where $\odot$ denotes the Hadamard product. By simultaneously maximizing the associations between $\bm{C}^{r^{(v)}}$ and both the clustering labels and the treatment feature, the indicator vector $\bm{e}^{r^{(v)}}$ can be adaptively optimized to extract the proxy confounders from $\bm{Z}^{r^{(v)}}$. Accordingly, Eq.~(\ref{M.2}) can be reformulated as follows:
\begin{equation}\label{M.3}
\begin{aligned}
\min_{\bm{\tau}^{(v)}\!,\bm{e}^{r^{(v)}}}& \!\|\!\frac{\sum_{i\in\nabla_{r}^{(v)}}\!\!\phi(\tau_{i}^{(v)} \!\bm{C}_{\cdot i}^{r^{(v)}}\!)}{|\nabla_{r}^{(v)}|} \!-\! \frac{\sum_{j\in\Delta_{r}^{(v)}}\!\!\phi(\tau_{j}^{(v)} \!\bm{C}_{\cdot j}^{r^{(v)}}\!)}{|\Delta_{r}^{(v)}|}\|_{\mathcal{H}_{\kappa}}^{2}\\
&+\|\bm{C}^{r^{(v)} \top}\bm{C}^{r^{(v)}}\!\!-\bm{F}\bm{F}^{\top}\|_{\mathrm{F}}^{2}-\operatorname{Tr}(\bm{C}^{r^{(v)} \top}\bm{P}^{r^{(v)}})\\
\text { s.t. } \tau_{i}^{(v)}& \geq 0, \bm{1}^{\top}\bm{\tau}^{(v)}=1, \bm{e}^{r^{(v)}} \in \{0,1\},
\end{aligned}
\end{equation}
where $\bm{P}^{r^{(v)}}\!\!\!\!=\!\!\textbf{1}_{d_v-1} \!\otimes\! \bm{X}_{r \cdot}^{(v)}$, and $\otimes$ denotes the Kronecker product.

Furthermore, to capture causal contributions of all features, a common strategy is to iteratively treat each view-specific feature as the treatment feature, while learning corresponding sample weights to mitigate confounding bias~\cite{ni2024feature,kuang2019treatment}. However, this strategy becomes computationally prohibitive in high-dimensional settings. Given that different views share the same set of samples, we propose learning view-shared sample weights to achieve confounding balance. Moreover, prior studies have demonstrated that mitigating confounding bias for a carefully selected subset of prototype features can help reduce spurious correlations across the entire feature set~\cite{wang2019blessings,vander2011new}. Accordingly, we propose balancing confounders for $m~(m \ll d_v)$ causally representative prototype features within each view, rather than all view-specific features. To this end, we introduce a causality-guided hierarchical feature clustering method to select $m$ prototype features from each view, with their indices denoted as $\Lambda^{(v)}$. The inter-feature distances are defined by the differences in their causal contributions as represented in  $\bm{W}^{(v)}$,  which is jointly learned during the confounder balancing process. We then learn view-shared sample weights to balance confounding distributions associated with the $m$ prototype features and formulate the proposed causal regularization module as follows:
\begin{equation}\label{M.4}
\begin{aligned}
\min_{\bm{\tau}\!,\bm{E}^{{(v)}} }& \!\sum_{v=1}^{V}\!\sum_{r=1}^{m}\!\|\!\frac{\sum_{i\in\nabla_{r}^{(v)}}\!\!\phi(\!\tau_{i} \bm{C}_{\cdot i}^{r^{(v)}}\!)}{|\nabla_{r}^{(v)}|} \!-\! \frac{\sum_{j\in\Delta_{r}^{(v)}}\!\!\phi(\!\tau_{j}\bm{C}_{\cdot j}^{r^{(v)}}\!)}{|\Delta_{r}^{(v)}|}\|_{\mathcal{H}_{\kappa}}^{2}\\
&+\|\bm{C}^{r^{(v)} \top}\bm{C}^{r^{(v)}}\!\!-\bm{F}\bm{F}^{\top}\|_{\mathrm{F}}^{2}-\operatorname{Tr}(\bm{C}^{r^{(v)} \top}\bm{P}^{r^{(v)}})\\
\text { s.t. } \tau&_{i} \geq 0, \bm{1}^{\top}\bm{\tau}=1, r \in \Lambda^{(v)}, \bm{e}^{r^{(v)}} \in \{0,1\}.
\end{aligned}
\end{equation}

By combining Eqs.~(\ref{M.1}) and (\ref{M.4}), we can obtain the final objective of the proposed CAUSA method as follows:
\begin{equation}\label{M.5}
\begin{aligned}
&\min_{\bm{\Theta}}\sum_{v=1}^{V} \{ \alpha\sum_{i=1}^{n}\tau_{i}\|\bm{X}^{(v) \top}_{\cdot i}\bm{W}^{(v)} - \bm{F}_{\cdot i} \|_{2}^{2}+\lambda \|\bm{W}^{(v)}\|_{2,1}\\
&+\!\beta\sum_{r=1}^{m}\|\frac{\sum_{i\in\nabla_{r}^{(v)}}\!\phi(\tau_{i} \bm{C}_{\cdot i}^{r^{(v)}})}{|\nabla_{r}^{(v)}|} \!-\! \frac{\sum_{j\in\Delta_{r}^{(v)}}\!\phi(\tau_{j}\bm{C}_{\cdot j}^{r^{(v)}})}{|\Delta_{r}^{(v)}|}\|_{\mathcal{H}_{\kappa}}^{2}\\
&+\!\|\bm{C}^{r^{(v)} \!\top}\!\bm{C}^{r^{(v)}}\!\!\!\!-\!\bm{F}\bm{F}^{\!\top}\!\|_{\mathrm{F}}^{2}\!-\!\!\operatorname{Tr}(\bm{C}^{r^{(v)} \!\top}\!\!\bm{P}^{r^{(v)}})\}\!+\!\operatorname{Tr}(\bm{F}^{\!\top}\!\bm{L}\bm{F})\\
&\text { s.t. } \bm{F}^{\top}\bm{F}\!=\!\bm{I}, \bm{F} \geq 0,\bm{W}^{(v) \top}\bm{W}^{(v)}\!=\!\bm{I},r \in \Lambda^{(v)},\tau_{i} \geq 0,\\
&\quad \,\,\,\, \bm{1}^{\top}\bm{\tau}=1,\bm{E}^{{(v)}} \in \{0,1\},
\end{aligned}
\end{equation}
where $\bm{\Theta}=\{\{\bm{W}^{(v)}\}_{v=1}^{V},\bm{F},\bm{\tau},\{\bm{E}^{{(v)}}\}_{v=1}^{V}\}$, $\bm{E}^{{(v)}} \in \mathbb{R}^{(d_v-1) \times m}$ denotes the confounding indicator matrix, with its $r$-th column given by $\bm{E}^{(v)}_{\cdot r} = \bm{e}^{r^{(v)}}$.

In summary, the proposed CAUSA in Eq.~(\ref{M.5}) offers the following advantages: (\romannumeral1) Unlike existing MUFS approaches that overlook the underlying causal mechanisms in the data,  CAUSA integrates feature selection and confounder balancing within a unified learning framework, thereby enabling the identification of causally informative features. (\romannumeral2) Rather than treating all observed features as confounders, CAUSA adaptively separates confounders from non-confounders to improve the balancing process.  (\romannumeral3) By learning view-shared sample weights and selecting causally representative prototype features, CAUSA achieves effective confounder balancing while reducing computational cost.

\section{Optimization and Analyses}
Since Eq. (\ref{M.5}) is not jointly convex in all variables, we propose an iterative optimization algorithm that alternately updates each variable while keeping the others fixed.

\noindent\textbf{Update $\bm{W}^{(v)}$ by Fixing Others.}
With other variables fixed, the objective function w.r.t. $\bm{W}^{(v)}$ can be transformed into:
\begin{equation}\label{O.1}
\begin{aligned}
&\min_{\bm{W}^{(v) \top}\bm{W}^{(v)}=\bm{I}}\operatorname{Tr}(\bm{W}^{(v) \top}\bm{A}^{(v)}\bm{W}^{(v)}-2\bm{W}^{(v) \top}\bm{B}^{(v)}),
\end{aligned}
\end{equation}
where $\bm{A}^{(v)}=\alpha\bm{X}^{(v)}\bm{\Gamma}\bm{X}^{(v) \top}+\lambda \bm{D}^{(v)}$, $\bm{\Gamma} = \operatorname{diag}(\bm{\tau})$, $\bm{D}^{(v)}$ is a diagonal matrix whose $i$-th diagonal entry is given by ${D}_{ii}^{(v)}= 1 / 2\sqrt{\|\bm{W}_{i \cdot}^{(v)}\|_{2}^{2}+\epsilon}$ ($\epsilon$ is a small constant to prevent the denominator from vanishing), and $\bm{B}^{(v)}=\alpha \bm{X}^{(v)}\bm{\Gamma}\bm{F}$. Problem (\ref{O.1}) can be efficiently solved using the generalized power iteration (GPI) algorithm~\cite{nie2017generalized}.

\noindent\textbf{Update $\bm{F}$ by Fixing Others.} With other variables fixed, we can formulate the Lagrangian function w.r.t. $\bm{F}$, and obtain the following optimal solution of $\bm{F}$ by applying the Karush–Kuhn–Tucker (KKT) conditions~\cite{boyd2004convex}:
\begin{equation}\label{O.3}
\begin{aligned}
\bm{F} \!=\! \bm{F} \!\odot\! \frac{\sum_{v=1}^{V}\!( \alpha \bm{J}_{[+]}^{(v)} \!+\!\bm{L}_{[-]}^{(v)}\bm{F}\!+\!2\bm{Q}^{(v)})  \!+\! 2\rho\bm{F}}{\sum_{v=1}^{V}\!(\alpha \bm{J}_{[-]}^{(v)}\!\!+\!\!\bm{L}_{[+]}^{(v)}\bm{F}) \!+\! \alpha V \bm{\Gamma}\!\bm{F}\!+\!\xi\bm{F}\!\bm{F}^{\!\top}\!\bm{F} },
\end{aligned}
\end{equation}
where $\bm{J}^{(v)}=\bm{\Gamma}\bm{X}^{(v) \top}\bm{W}^{(v)}$, $\bm{Q}^{(v)}=\sum_{r=1}^{m}\bm{C}^{r^{(v)} \!\top}\!\bm{C}^{r^{(v)}}\!\bm{F}$, $\xi\!=\!2(mV+\rho)$, and $\rho$ is a large constant used to ensure the orthogonality of $\bm{F}$. Moreover, for any matrix $\bm{M}$, $\bm{M}_{[+]}$ and $\bm{M}_{[-]}$ are respectively defined by $\bm{M}_{[+]}=1/2(|\bm{M}|+\bm{M})$ and $\bm{M}_{[-]}=1/2(|\bm{M}|-\bm{M})$.

\noindent\textbf{Update $\bm{E}^{(v)}$ by Fixing Others.} Following commonly adopted relaxation strategies for discrete constraints~\cite{xing2021fairness}, we relax $\bm{E}^{(v)}$ to take continuous values in the range $[0,1]$, thereby reformulating the optimization problem w.r.t. $\bm{E}^{(v)}$ as follows:
\begin{equation}\label{O.5}
\begin{aligned}
\min_{\bm{E}_{\cdot r}^{(v)}} & \frac{\beta}{n^2}\!\sum_{v=1}^{V}\sum_{i,j=1}^{n}\! \tau_{i}\tau_{j}K_{ij}^{r^{(v)}}\bm{C}_{\cdot i}^{r^{(v)}\top}\!\bm{C}_{\cdot j}^{r^{(v)}}\!\!-\!\operatorname{Tr}(\bm{C}^{r^{(v)} \top}\!\bm{P}^{r^{(v)}})  \\
&+\!\|\bm{C}^{r^{(v)} \top}\!\bm{C}^{r^{(v)}}\!\!\!\!-\!\bm{F}\bm{F}^{\!\top}\!\|_{\mathrm{F}}^{2} + \varrho\|\bm{E}_{\cdot r}^{(v)}\|_{1}\\
\text { s.t. } &\bm{E}_{\cdot r}^{(v)} \in [0,1],
\end{aligned}
\end{equation}
where $\varrho$ is a large constant used to enforce sparsity of $\bm{E}_{\cdot r}^{(v)}$. Problem (\ref{O.5}) can be efficiently solved using the proximal gradient descent method~\cite{nitanda2014stochastic}.

\noindent\textbf{Update $\bm{\tau}$ by Fixing Others.} Based on the empirical estimation of MMD from~\cite{kumagai2019unsupervised}, the optimization problem w.r.t. $\bm{\tau}$ can be reformulated as follows:
\begin{equation}\label{O.4}
\begin{aligned}
&\min_{\bm{\tau}} \frac{\beta}{n^2}\bm{\tau}^{\top}\bm{H}\bm{\tau} + \alpha \bm{\tau}^{\top}\bm{g}   \\
&\text { s.t. } \tau_{i} \geq 0, \bm{1}^{\top}\bm{\tau}=1,
\end{aligned}
\end{equation}
where $H_{ij}\!=\!\sum_{v=1}^{V}\sum_{r=1}^{m}K_{ij}^{r^{(v)}}\bm{C}_{\cdot i}^{r^{(v)}\top}\!\bm{C}_{\cdot j}^{r^{(v)}}$, $K_{ij}^{r^{(v)}}\!=\!\Omega_{ri}^{(v)}\Omega_{rj}^{(v)}\!-\!2\Omega_{ri}^{(v)}(1\!-\!\Omega_{rj}^{(v)})+(1\!-\!\Omega_{ri}^{(v)})(1\!-\!\Omega_{rj}^{(v)})$, and $\Omega_{ri}^{(v)}$ denotes an indicator variable for view $v$ which equals 1 if the $i$-th sample is assigned to the treatment group based on the value of the $r$-th treatment feature, and 0 otherwise. Moreover, $\bm{g}$ is a vector whose $i$-th entry is given by $g_i=\sum_{v=1}^{V}\|\bm{X}_{\cdot i}^{(v) \top}\bm{W}^{(v)}-\bm{F}_{i \cdot}\|_{2}$, where $\|\cdot\|_{2}$ denotes the $\ell_{2}$-norm. Problem (\ref{O.4}) is a standard simplex projection problem, which can be efficiently solved using the method proposed in~\cite{duchi2008efficient}.

\begin{algorithm}[t]
\caption{Iterative Algorithm of CAUSA}
\textbf{Input:} \makebox[0pt][l]{Multi-view data $\{\bm{X}^{(v)}\}_{v=1}^{V}$; the parameters $\alpha$, $\beta$,}\\
\makebox[0pt][l]{\hspace*{3.2em}and $\lambda$; the number of prototype features $m$.}

\begin{algorithmic}[1]
    \STATE {Initialize} $\{\bm{W}^{(v)}\}_{v=1}^{V}$, $\{\bm{E}^{(v)}\}_{v=1}^{V}$, $\bm{F}$, $\bm{\tau}$.

    \WHILE{not convergent}

    \STATE Update $\{\bm{W}^{(v)}\}_{v=1}^{V}$ by solving Eq. (\ref{O.1});

    \STATE Update $\bm{F}$ via Eq. (\ref{O.3});

    \STATE Update $\{\bm{E}^{(v)}\}_{v=1}^{V}$ by solving Eq. (\ref{O.5});
    
    \STATE Update $\bm{\tau}$  by solving Eq. (\ref{O.4});

    \STATE Update $\{\Lambda^{(v)}\}_{v=1}^{V}$ using causally-guided hierarchical feature clustering;
    
    \ENDWHILE
\end{algorithmic}
\textbf{Output:} {Sorting the $\ell_{2}$-norm of rows of $\{\bm{W}^{(v)} \}_{v=1}^{V}$ in descending order and selecting the top $h$ features.}
\end{algorithm}
Algorithm 1 summarizes the optimization process of the proposed CAUSA. In the initialization step, $\bm{W}^{(v)}$ and $\bm{E}^{(v)}$ are set as matrices with all ones, $\bm{F}$ is initialized using spectral clustering, and $\tau_i$ is set to $1/n$ for all $i$. Moreover, $m$ is empirically set to 15.

\noindent\textbf{Time Complexity Analysis.} In each iteration of Algorithm 1, updating $\bm{W}^{(v)}$ using the GPI algorithm costs $\mathcal{O}(d_{v}^{2}c)$. The
computational cost for updating $\bm{F}$ and $\tau$ is $\mathcal{O}(n^2dm)$, where $d = \sum_{v=1}^{V} d_{v}$. Updating $\bm{E}^{(v)}$ using the proximal gradient descent method requires $\mathcal{O}(n^{2}d_{v}m)$. Thus, the overall time complexity of Algorithm 1 is $\mathcal{O}(n^{2}dm+\sum_{v=1}^{V}d_{v}^{2}c)$.

\section{Experiments}
\subsection{Experimental Settings}
\noindent\textbf{Datasets.} We conduct experiments on six real-world multi-view datasets, including MSRA~\cite{zhang2017latent}, MSRC~\cite{huang2022multi}, Prokaryotic~\cite{wen2023unpaired}, SensIT~\cite{chang2011libsvm}, CCV~\cite{wang2019multi}, and FMNIST~\cite{xiao2017fashion}. Table~\ref{Data} summarizes the statistical details of these datasets.

\noindent\textbf{Compared Methods.} We compare the proposed CAUSA with several SOTA methods, including SPDFS~\cite{SPDFS}, BDGFS~\cite{BDGFS}, GCDUFS~\cite{GCDUFS}, UKMFS~\cite{UKMFS}, WLTL~\cite{WLTL}, JMVFG~\cite{JMVFG}, OEDFS~\cite{OEDFS}, CvLP-DCL~\cite{CvLPDCL}, as well as a baseline (AllFea) that utilizes all original features.

\noindent\textbf{Comparison Schemes.} To ensure a fair comparison, we employ the grid search strategy to tune the parameters of all methods and report their respective optimal results. Simultaneously, the parameters $\alpha$, $\beta$ and $\lambda$ of our method are tuned over the range $\{10^{-3}, 10^{-2}, 10^{-1}, 1, 10^{1}, 10^{2}, 10^{3}\}$. Because it is challenging to determine the optimal number of features~\cite{li2017feature}, we vary the feature selection ratios from 10\% to 50\% in 10\% increments across all datasets. Following a common practice for assessing MUFS~\cite{FSDK,hou2023adaptive}, we employ two widely used clustering metrics, Clustering Accuracy (ACC) and Normalized Mutual Information (NMI), to evaluate the quality of selected features. We run K-means clustering 50 times on the selected features and report the average result. All experiments were conducted using MATLAB R2022b on a desktop equipped with an Intel Core i9-10900 CPU (2.80 GHz) and 64 GB of RAM.




			


\begin{table}[t]
\centering
\small
\caption{Statistics of different datasets}\label{Data}
\setlength{\tabcolsep}{0.25mm}
        \begin{tabular}{@{\extracolsep{\fill}}lcccc}
            \toprule
            Datasets  & Views & Samples & Features & Classes \\
            \midrule
            MSRA      & 6 & 210 & 1302/48/512/100/256/210 & 7 \\
            MSRC      & 5 & 210 & 24/576/512/256/254 & 7 \\
            Prokaryotic & 2 & 551 & 393/438 & 4 \\
            SensIT    & 2 & 1500 & 50/50 & 3 \\
            CCV       & 3 & 6773 & 20/20/20 & 20 \\
            FMNIST    & 3 & 8000 & 512/512/1280 & 10 \\
            \toprule
        \end{tabular}
\end{table}

\begin{figure}[t]
\centering
\includegraphics[width=0.95\columnwidth]{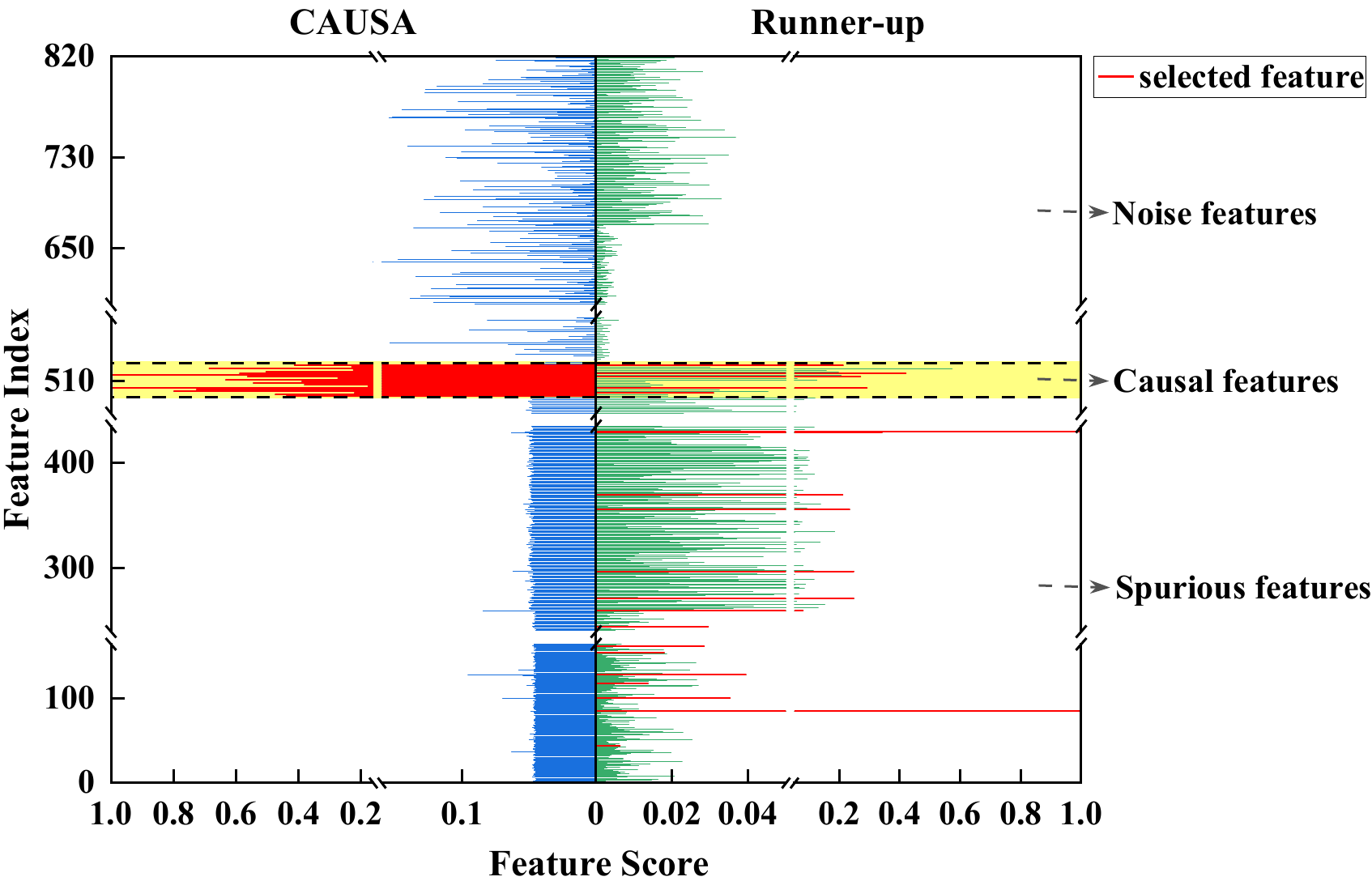} 
\caption{Feature selection results of CAUSA and the runner-up method on the synthetic dataset. Yellow area: causal feature indices; others: non-causal feature indices.}
\label{Simu}
\end{figure}

\begin{figure}[t]
\centering
\includegraphics[width=\columnwidth]{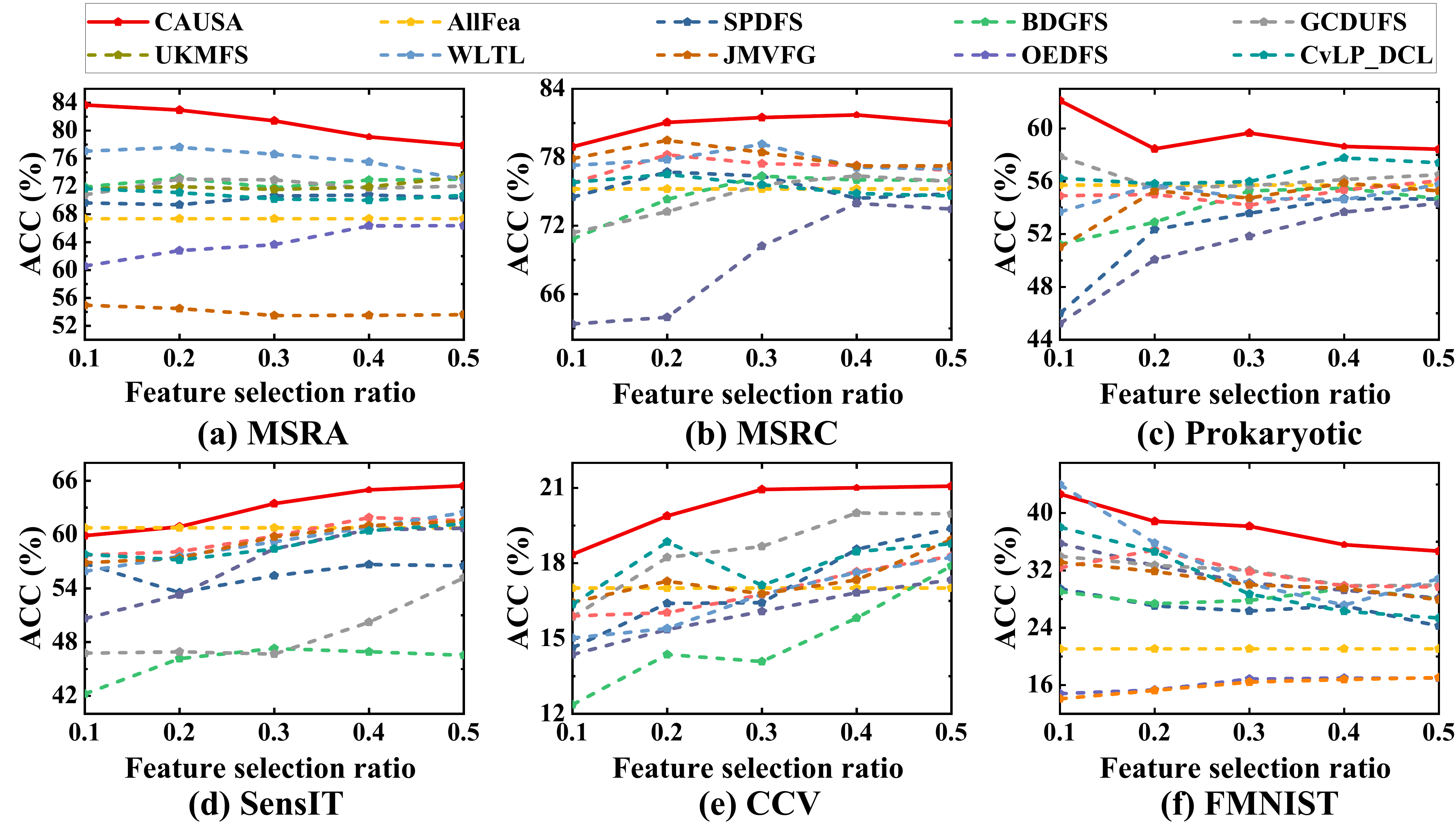} 
\caption{ACC of different methods on six multi-view datasets under different feature selection ratios.}
\label{ACC}
\end{figure}

\begin{table*}[t]
\centering
\small
\caption{The average ACC and NMI values (\%) of different methods on six multi-view datasets, where boldface indicates the best performance and $*$ denotes significant improvements according to the Wilcoxon signed-rank test.}\label{Clustering}
\setlength{\tabcolsep}{2.2mm}
\begin{tabular}{lcccccccccccc}
\toprule
\multirow{2}*{Methods} & \multicolumn{2}{c}{MSRA}&\multicolumn{2}{c}{MSRC} &\multicolumn{2}{c}{Prokaryotic} &\multicolumn{2}{c}{SensIT}&\multicolumn{2}{c}{CCV} &\multicolumn{2}{c}{FMNIST}\\
\cmidrule(r){2-3} \cmidrule(r){4-5} \cmidrule(r){6-7} \cmidrule(r){8-9} \cmidrule(r){10-11} \cmidrule(r){12-13} 
~&ACC&NMI&ACC&NMI&ACC&NMI&ACC&NMI&ACC&NMI&ACC&NMI\\
\midrule
{CAUSA} & \textbf{81.40}$^{*}$&  \textbf{76.52}$^{*}$&  \textbf{81.48}$^{*}$&  \textbf{74.56}$^{*}$&  \textbf{59.65}$^{*}$&  \textbf{24.20}$^{*}$&  \textbf{63.45}$^{*}$&  \textbf{22.62}$^{*}$&  \textbf{20.94}$^{*}$&  \textbf{19.23}$^{*}$&  \textbf{38.15}$^{*}$&  \textbf{32.29}$^{*}$\\

{AllFea}  &67.37& 63.27& 75.22& 66.80& 55.71& 2.68& 60.76& 20.53& 17.01& 14.10& 21.05& 11.10  \\

{SPDFS}  &70.64& 64.96& 76.35& 69.58& 53.57& 0.71& 55.40& 14.59& 16.43& 11.35& 26.33& 16.68 \\

{BDGFS}  &71.82& 65.30& 76.33& 68.16& 55.28& 7.51& 47.29& 9.21& 14.08& 9.26& 27.80& 19.42 \\

{GCDUFS}  &72.93& 68.60& 75.61& 67.56& 55.65& 10.99& 46.66& 9.28& 18.67& 15.42& 32.01& 23.95 \\

{UKMFS}  &71.57& 67.40& 77.43& 69.90& 54.19& 16.41& 59.82& 17.50& 16.75& 13.73& 31.83& 22.90 \\

{WLTL}  &76.61& 71.70& 79.14& 71.72& 54.69& 9.02& 59.19& 17.77& 16.76& 13.71& 30.12& 23.97 \\

{JMVFG}  &53.48& 69.37& 78.43& 71.44& 54.73& 13.86& 59.75& 17.92& 16.78& 13.83& 30.01& 21.79 \\

{OEDFS}  &63.64& 55.82& 70.22& 62.46& 51.83& 1.30& 58.43& 16.28& 16.08& 13.26& 30.25& 24.49 \\

{CvLP-DCL}  &70.19& 62.77& 75.59& 66.68& 55.97& 14.46& 58.41& 16.74& 17.12& 14.65& 28.69& 20.41 \\
\bottomrule
\end{tabular}
\end{table*}

\begin{figure}[t]
\centering
\includegraphics[width=\columnwidth]{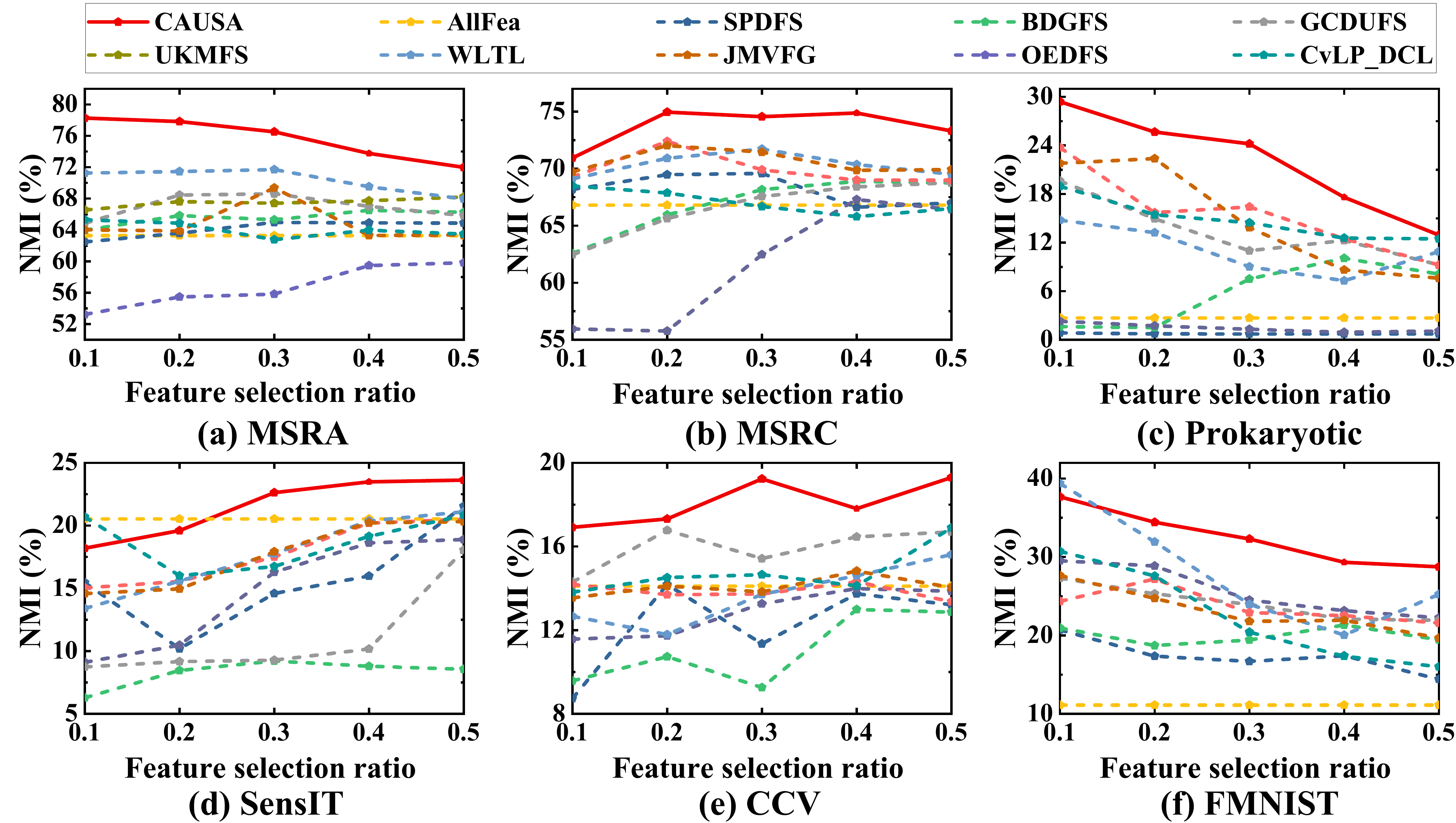} 
\caption{NMI of different methods on six multi-view datasets under different feature selection ratios.}
\label{NMI}
\end{figure}

\begin{figure}[t]
\centering
\includegraphics[width=\columnwidth]{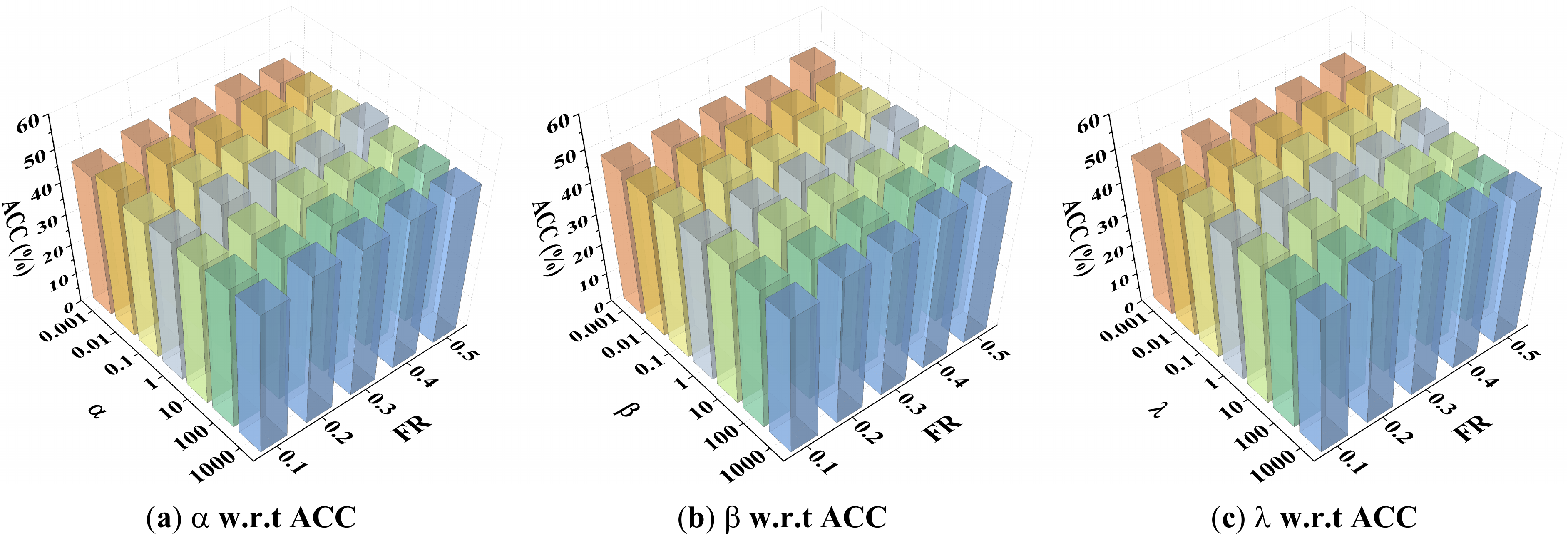} 
\caption{ACC of CAUSA with varying parameters $\alpha$, $\beta$, $\lambda$ and feature selection ratios on MSRA dataset.}
\label{Sensitivity}
\end{figure}

\begin{figure}[t]
\centering
\includegraphics[width=\columnwidth]{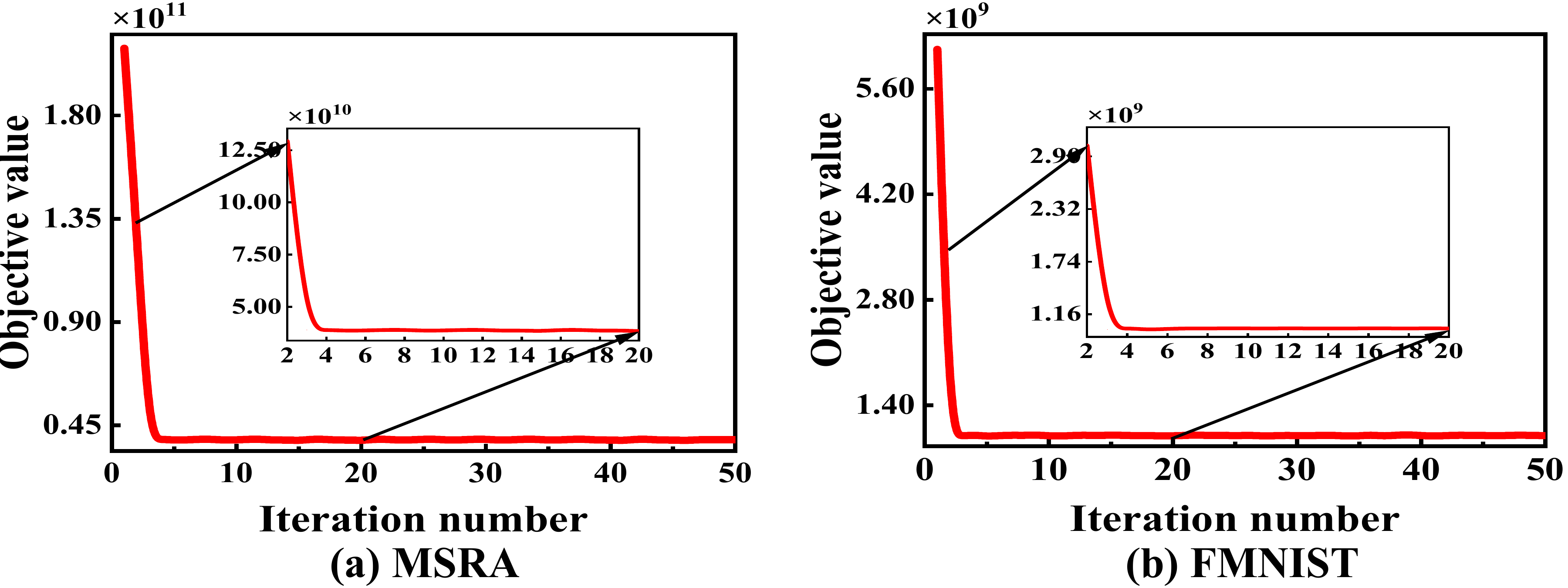} 
\caption{Convergence curves of CAUSA on MSRA and FMNIST datasets.}
\label{Convergence}
\end{figure}

\begin{figure}[t]
\centering
\includegraphics[width=\columnwidth]{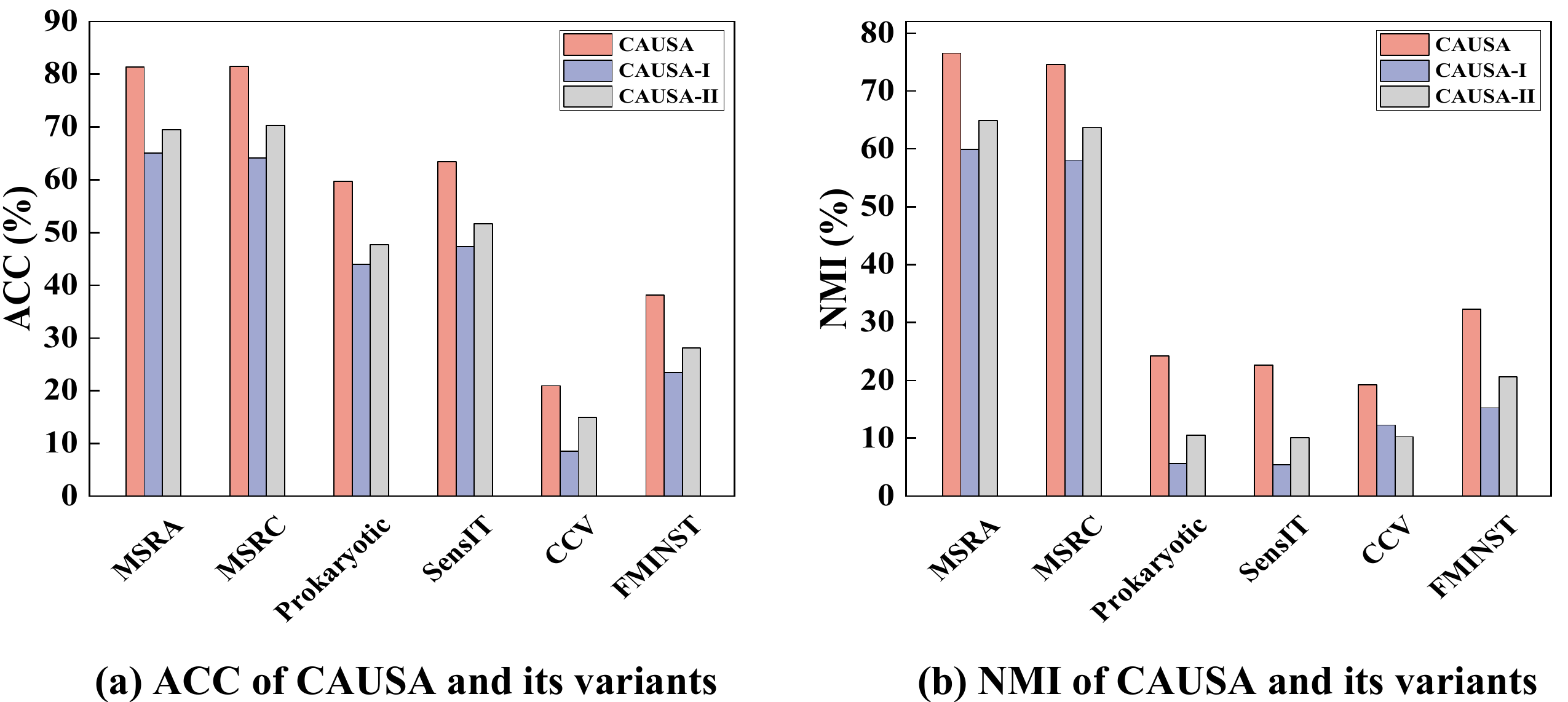} 
\caption{Performance comparison of CAUSA and its two variants in terms of ACC and NMI.}
\label{Ablation}
\end{figure}

\subsection{Experiments on Synthetic Dataset}
Following common data generation protocols in causal literature~\cite{lu2021invariant,kallus2018causal}, we construct a synthetic multi-view dataset to evaluate the effectiveness of CAUSA in mitigating spurious correlations and identifying causal features. The synthetic dataset consists of 500 samples with two views, each view containing 10 causal features. In addition, the first and second views contain 410 and 390 non-causal features, respectively, of which 260 and 240 are spurious features and 150 are noise features. The causal features of each view are independently sampled from Gaussian distributions and directly determine the generation of class labels, thereby establishing causal dependencies between the features and the labels. Spurious features are generated from confounding factors, yielding apparent but non-causal correlations with the labels. Noise features are randomly sampled from a Gaussian distribution and have negligible correlations with the labels. Fig.~\ref{Simu} illustrates the feature selection results of CAUSA and the runner-up method on the synthetic dataset. The experimental results indicate that CAUSA effectively identifies causal features, while the runner-up method frequently selects non-causal features with spurious correlations.

\subsection{Experiments on Real Multi-view Datasets}
\noindent\textbf{Performance Comparisons.} Table~\ref{Clustering} presents the ACC and NMI results for CAUSA and competing methods when 30\% of features are selected. The results show that CAUSA consistently outperforms other competing methods across all datasets in both ACC and NMI. For MSRA and FMNIST datasets, CAUSA gains over 9\% and 11\% average improvement in terms of ACC and NMI, respectively. On Prokaryotic dataset, CAUSA yields improvements of more than 5\% in ACC and 15\% in NMI on average. As to MSRC, SensIT, and CCV datasets, CAUSA achieves average improvements exceeding 4\% in ACC and 5\% in NMI. Furthermore, to provide a comprehensive evaluation of CAUSA, we report the performance of all methods under varying feature selection ratios, as shown in Figs.~\ref{ACC} and~\ref{NMI}. We can observe that CAUSA achieves superior performance over other competing methods in most cases when the feature selection ratio ranges from 10\% to 50\%. The superior performance of CAUSA is attributed to adaptively separating confounding features from multiple views, and jointly learning feature selection and confounder balancing, thereby enabling the identification of causally informative features.

\noindent\textbf{Parameter Sensitivity and Convergence Analysis.} To investigate the sensitivity of CAUSA to $\alpha$, $\beta$, and $\lambda$, Fig.~\ref{Sensitivity} shows how its performance varies as these parameters and the feature selection ratio (FR) change. We can observe that CAUSA performs relatively stably w.r.t. all parameters. Moreover, Fig.~\ref{Convergence} shows that the proposed optimization algorithm quickly reduces the objective function value and converges within 20 iterations.

\noindent\textbf{Ablation Study.}
This ablation study investigates the effectiveness of each component in  CAUSA by comparing it to its two variants: (\romannumeral1) CAUSA-\uppercase\expandafter{\romannumeral1}: with the causal regularization module removed from Eq.~(\ref{M.5}); (\romannumeral2) CAUSA-\uppercase\expandafter{\romannumeral2}: with the adaptive confounder separation removed from Eq.~(\ref{M.5}) and all observed features treated as confounders. Fig.~\ref{Ablation} presents the ablation results across six datasets. Both CAUSA-\uppercase\expandafter{\romannumeral1} and CAUSA-\uppercase\expandafter{\romannumeral2} show a substantial reduction in ACC and NMI performance relative to CAUSA. These results highlight the effectiveness of jointly learning MUFS and confounder balancing for identifying causal informative features, as well as the efficacy of adaptive confounder separation.

\section{Conclusion}
In this paper, we analyze MUFS from a causal perspective and find that existing methods may select irrelevant features due to their ignoring spurious correlations induced by confounders. To identify causally informative features, we propose a novel MUFS method, termed CAUSA. CAUSA first embeds MUFS into a generalized unsupervised spectral regression model, and then introduces a causal regulation module that adaptively separates confounders from multiple views and learns view-shared sample weights to balance confounder distributions. By jointly learning MUFS and confounder balancing, CAUSA effectively mitigates spurious correlations and selects causally informative features. Comprehensive experiments demonstrate the superiority of CAUSA over SOTA methods.

\bibliographystyle{IEEEtran}
\bibliography{IEEEabrv}

\end{document}